\pdfoutput=1

\documentclass[11pt]{article}
\usepackage{enumitem}

\usepackage[final]{acl}

\usepackage{times}
\usepackage{latexsym}

\usepackage{amsmath}

\usepackage{float}
 
\usepackage[bidi=default]{babel}
\usepackage[LAE, T1]{fontenc}

\usepackage[most]{tcolorbox}
\usepackage{xcolor}

\usepackage[T1]{fontenc}

\usepackage[utf8]{inputenc}

\usepackage{microtype}

\usepackage{multirow}
\usepackage{inconsolata}

\usepackage{graphicx}

\usepackage{xcolor}
\usepackage{mdframed}
\usepackage{listings} 
\usepackage{textcomp} 
\usepackage{caption}
\usepackage{booktabs}
\newcommand{\taha}[1]{\textcolor{black}{#1}}
\newcommand{\asim}[1]{\textcolor{black}{#1}}

\mdfsetup{
    innertopmargin=8pt,    
    innerbottommargin=8pt, 
    innerleftmargin=8pt,   
    innerrightmargin=8pt,  
    backgroundcolor=gray!15, 
    linewidth=0.5pt,       
    linecolor=black        
}

\title{Tool Calling for Arabic LLMs: Data Strategies and Instruction Tuning}

\author{Author 1 \and ... \and Author n \\
        Address line \\ ... \\ Address line}

\author{
  Asim Ersoy,
  Enes Altinisik,
  Husrev Taha Sencar,
  Kareem Darwish 
  \\
  \\
  Qatar Computing Research Institute, HBKU, Qatar \\
  \small{\{aersoy,ealtinisik,hsencar,kdarwish\}@hbku.edu.qa}
}

\begin{document}
\maketitle
\begin{abstract}
Tool calling is a critical capability that allows Large Language Models (LLMs) to interact with external systems, significantly expanding their utility. However, research and resources for tool calling are predominantly English-centric, leaving a gap in our understanding of how to enable this functionality for other languages, such as Arabic. This paper investigates three key research questions: (1) the necessity of in-language (Arabic) tool-calling data versus relying on cross-lingual transfer, (2) the effect of general-purpose instruction tuning on tool-calling performance, and (3) the value of fine-tuning on specific, high-priority tools. To address these questions, we conduct extensive experiments using base and post-trained variants of an open-weight Arabic LLM. To enable this study, we bridge the resource gap by translating and adapting two open-source tool-calling datasets into Arabic. Our findings provide crucial insights into the optimal strategies for developing robust tool-augmented agents for Arabic.
\end{abstract}

\section{Introduction}
Tool calling, frequently referred to as function calling, represents a pivotal feature that significantly extends the operational capabilities of Large Language Models (LLMs) and LLM-based agents. This functionality empowers an LLM to interact with external systems or applications by generating structured requests in response to a user's natural language prompt \cite{masterman2024landscape}, allowing the LLMs to perform tasks beyond their intrinsic capabilities. Typically, an LLM is provided with a prompt alongside a predefined set of tools (or functions), complete with their descriptions, arguments, and expected output. The LLM then analyzes the prompt to determine if invoking an external tool is necessary to fulfill a user's request. If a tool call is identified, the LLM generates a structured tool call request, in accordance with what the tools expect. The output generated by the execution of the external tool is subsequently fed back to the LLM to be incorporated into the final response of the LLM, thereby creating a dynamic and iterative problem-solving loop \cite{masterman2024landscape}. Consequently, an LLM must be explicitly trained to understand tool descriptions, recognize when they are needed, generate structured function calls, and handle their output. Figure \ref{box:example} shows an example addition function with its invocation. 

\begin{mdframed}
\small
\label{tool_usage_example} 

\begin{lstlisting}
def add(a:float, b:float) -> float:
    """
    Add two numbers together.
    Args:
        a: First number to add
        b: Second number to add
    Returns:
        The sum of the two numbers
    """
    return a + b
\end{lstlisting}
\centering{\textbf{(a)} Addition tool} \\
\vspace{8pt} 
\hrule       
\vspace{8pt}
\begin{lstlisting}
What is the sum of 256 and 67?
\end{lstlisting}
\centering{\textbf{(b)} Example prompt to trigger tool call} \\
\vspace{8pt}
\hrule
\vspace{8pt}


\begin{lstlisting}
{"name": "add", "arguments": {"a": 256, "b": 67}}
\end{lstlisting}
\centering{\textbf{(c)} LLM generated function call} \\
\vspace{8pt}
\hrule
\vspace{8pt}

\begin{lstlisting}
313
\end{lstlisting} 
\centering{(d) Tool output}

\vspace{8pt}
\hrule
\vspace{8pt}

\begin{lstlisting}
The sum of 256 and 57 is 313.    
\end{lstlisting}
\centering{(e) Final LLM response}
\end{mdframed}
\vspace{-6pt}

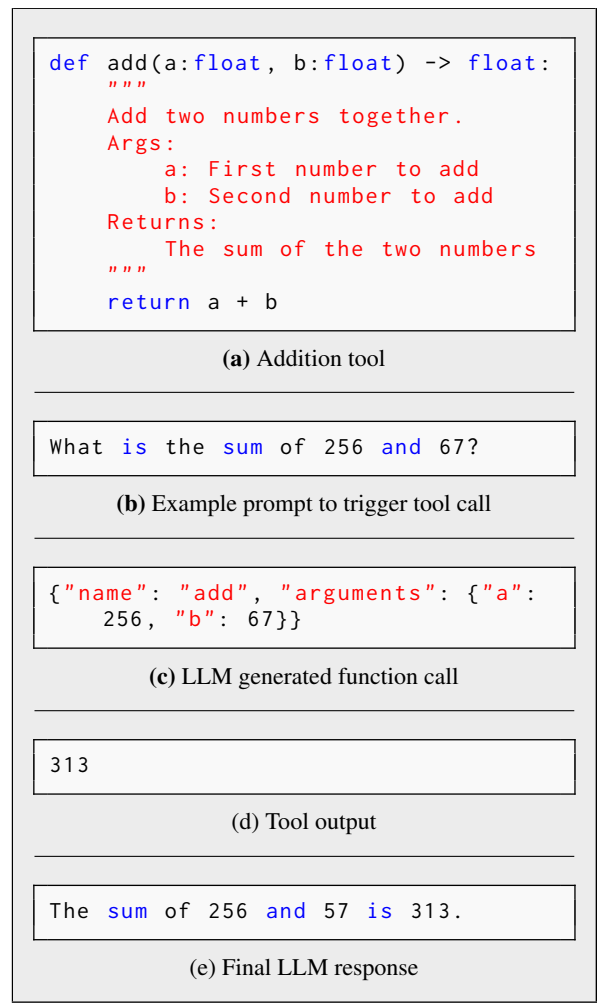
\captionof{figure}{Example tool with invocation}\label{box:example}

Currently, there are quite a few tool-calling datasets, such as Glaive\footnote{\url{https://huggingface.co/datasets/glaiveai/glaive-function-calling-v2}}, xLAM \cite{Liu2024APIGen}, ToolAce \cite{liu2024toolacewinningpointsllm}, and Hermes\footnote{\url{https://huggingface.co/datasets/NousResearch/hermes-function-calling-v1}}, that provide tens of thousands of different tools with the intended interaction with them.  The vast majority of tool-calling datasets are in English, with some that have been translated to other languages (e.g. Chinese Glaive\footnote{\url{https://huggingface.co/datasets/llamafactory/glaive_toolcall_zh}}). However, given the cross-lingual generalization capabilities of LLMs, it is not clear how much impact non-English training data has on the tool-calling abilities of the LLMs.  Further, though tool-calling training data demonstrate to an LLM all the required tool-calling steps, would tool-calling benefit from LLM supervised fine-tuning (SFT) on general-purpose tasks such as chat, summarization, or headline generation?  Additionally, tool-calling training data can't cover all possible tools, and LLMs are expected to generalize to new tools. However, if a set of tools is important to a user or an organization, how much benefit would be observed if training samples for these specific calls are included in tool-calling training data?


This paper attempts to answer the three aforementioned research questions, namely:
\begin{enumerate}[leftmargin=*]
    \item When using tool-calling for non-English prompts (e.g., Arabic), do LLMs benefit from being fine-tuned on tool-calling datasets in that language?
    \item What effect does post-training on general domain capability data have on the ability of LLMs to perform effective tool calling?
    \item Though LLMs can generalize well beyond the examples in their training data, is there value for tool-specific fine-tuning? 
\end{enumerate}

The contributions of the paper are as follows:
\begin{itemize}[leftmargin=*]
    \item We conduct extensive experimentation on a public open-weights LLM, Fanar \cite{fanarllm2025} that is specifically trained for Arabic, to answer the above research questions.
    \item We contribute a large dataset of Arabic tool-calling training set composed of tens of thousands of examples, and a version of the Fanar open-weight Arabic LLM that is fine-tuned for tool-calling.\footnote{\url{https://huggingface.co/collections/QCRI/arabictoolcalling-68b82e0b8f0865d6e3b179e7}}
\end{itemize}



\section{Related Work}
Tool calling relies on a model’s ability to detect user intent, decide when to invoke a tool, and translate the query into structured parameters aligned with the tool’s schema.
This process entails selecting the appropriate tool, adhering to its specification, extracting and formatting the input arguments, and generating responses that conform to the expected output format.
To improve LLM performance in tool use, several works have built instruction-tuning datasets that expose models to a diverse set of tools and usage patterns across varied prompt scenarios and interaction contexts \cite{Qin2024ToolLLM, Patil2024Gorilla, Liu2024APIGen, Abdelaziz2024GraniteFunction, liu2024toolacewinningpointsllm}. 

A key emphasis in these datasets is the breadth and complexity of tool coverage, with some efforts incorporating tens of thousands of real-world APIs spanning hundreds of domains \cite{Qin2024ToolLLM, liu2024toolacewinningpointsllm}. Beyond API diversity, these datasets increasingly capture advanced usage scenarios, including parallel and dependent tool invocations \cite{Liu2024APIGen, Abdelaziz2024GraniteFunction}, support for nested and structured parameter types \cite{liu2024toolacewinningpointsllm}, and multi-turn interactions that require contextual memory and dialogue state tracking \cite{Tang2023ToolAlpaca, Liu2024APIGen}. 
Additionally, several datasets aim to strengthen the planning and reasoning abilities required for effective tool use \cite{Huang2024UltraTool, Tang2023ToolAlpaca, ApiBank2024}.
While these datasets have advanced the tool-use capabilities of LLMs, an important open question is whether they enable sufficient generalization to non-English prompts and unseen domain-specific tools. We investigate this in the context of Arabic-language tool-use with a focus on a small set of real-world tools developed for deployment in culturally and linguistically specialized settings.

A more subtle and critical challenge is integrating tool use in a way that aligns with the model’s internal reasoning capabilities.
Ideally, a model should invoke a tool only when its own knowledge or inference abilities are insufficient to complete the task described in the user query, or when a tool is capable of performing a required step with greater efficiency and effectiveness. In this sense, effective tool use should be selective and autonomous, minimizing unnecessary calls and the associated computational or latency costs \cite{Chen2024ToolUseAlignment}. Achieving this balance requires careful design of the supervised fine-tuning and preference optimization stages, ensuring that general capabilities are calibrated to support—rather than compete with—tooling. 

\section{Datasets}

To construct our training and evaluation data, we utilized four distinct datasets (shown in Table \ref{tab:datasets}). We adapted two prominent open-source function-calling datasets, namely \texttt{Glaive}\footnote{\url{https://huggingface.co/datasets/glaiveai/glaive-function-calling-v2}} and \texttt{xLAM} \cite{Liu2024APIGen}, where we translated them into Arabic using Gemini-2.5-Flash-no-thinking \cite{team2023gemini} following the prompt templates described in Appendix \ref{sec:prompts}. In our experiment, we use the Arabic and English versions of the datasets in isolation or in combination.  We split both datasets into training and test splits, where the English and Arabic train and test splits are direct translations of each other.

To address specific use cases, we curated two novel datasets. \taha{The first, \texttt{CustomTools}, is a collection of unique tools synthetically generated using, again, Gemini. It includes both positive examples, where a function call is required, and negative examples, where a function call is not required or not present in the list of provided tools.  We synthesized Arabic and English examples. The tools cover functions such as translation, image generation, speech generation, speech recognition, text diacritization, Islamic knowledge, recent news, and person biography lookup. We list the function definitions in Appendix \ref{sec:customToolsDef}.}

The second, \texttt{IslamicRAGTool}, was built from real question-answer pairs obtained from the Fanar Arabic and English Islamic question-answering service API\footnote{\url{https://api.fanar.qa/docs}}. \texttt{IslamicRAGTool} is different from the other calls in three ways, namely: the dataset is based on actual logs instead of being synthetic; it involves specific topic/genre classification; and, unlike the other tools the LLM needs to pass either the user input or sequence of interactions as is without argument extraction. We provide a comprehensive overview of the datasets and their statistical properties in Table \ref{tab:datasets}, and examples from the datasets in Appendix \ref{sec:datasetExamples}.

\begin{table}[h!]
\centering
\caption{Summary of Function-Calling Datasets. Language denotes the language of the dataset (AR = Arabic, EN = English). FC indicates whether the examples include function calls (Y = Yes, N = No). Turns specifies whether interactions are single-turn (S) or multi-turn (M), while Calls denote whether a single (S) or multiple (M) function calls occur per turn. The Train and Test columns report the number of samples in each split. The datasets \texttt{Glaive, xLAM}, \texttt{CustomTools}, and \texttt{IslamicRAGTool} contain 972, 3,179, 8, and 1 unique tools, respectively, distributed across their examples.}
\label{tab:datasets}
\scriptsize 
\setlength{\tabcolsep}{3pt} 
\begin{tabular}{@{}lccccrrr@{}} 
\toprule
\textbf{Dataset} & \textbf{Language} & \textbf{FC} & \textbf{Turns} & \textbf{Calls} & \textbf{Train} & \textbf{Test} \\
\midrule
\multirow{4}{*}{Glaive}          & AR & Y & M & S & 37,684 & 1,953 \\
                & AR & N & M & S & 38,678 & 1,000 \\
                & EN & Y & M & S & 37,684 & 1,953 \\
                & EN & N & M & S & 38,678 & 1,000 \\
\midrule
\multirow{4}{*}{xLAM}            & AR & Y & S & M & 58,999 & 1,001 \\
                & AR & N & S & M & 19,361 & 1,077 \\
                & EN & Y & S & M & 58,999 & 1,001 \\
                & EN & N & S & M & 19,361 & 1,077 \\
\midrule
\multirow{4}{*}{CustomTools}      & AR & Y & S & S & 4,528  & 1,000 \\
                & AR & N & S & S & 4,313  & 1,000 \\
                & EN & Y & S & S & 5,133  & 1,000 \\
                & EN & N & S & S & 5,983  & 1,000 \\
\midrule
\multirow{4}{*}{IslamicRAGTool} & AR & Y & S & S & 10,000 & 1,000 \\
                & AR & N & S & S & 10,000 & 1,000 \\
                & EN & Y & S & S & 10,000 & 1,000 \\
                & EN & N & S & S & 10,000 & 1,000 \\
\bottomrule
\end{tabular}
\end{table}

\section{Experimental Design}

\taha{
\subsection{Experiments}
We designed five experiments to answer the three main research questions introduced in Section~1. Each experiment evaluates a different configuration of supervised fine-tuning and tool-calling training strategies.
\begin{itemize}[leftmargin=*]
\item \textbf{Experiment 1:} Fine-tuning of the \textbf{base} Fanar model using English tool-calling data drawn from a combination of the \texttt{Glaive} and \texttt{XLAM} datasets.
\item \textbf{Experiment 2:} A direct replication of Experiment~1, but using the translated Arabic versions of the tool-calling examples from \texttt{Glaive} and \texttt{XLAM}.
\item \textbf{Experiment 3:} Continued fine-tuning of \textbf{instruction-tuned} Fanar 
using a mix of English tool-calling examples from \texttt{Glaive} and \texttt{XLAM}.
\item \textbf{Experiment 4:} Similar to Experiment~3, but using bilingual tool-calling data (English and Arabic) from the \texttt{Glaive} and \texttt{XLAM} datasets.
\item \textbf{Experiment 5:} Similar to Experiment~4, where we fine-tuned the instruction tuned Fanar model with the bilingual training sets of of \texttt{Glaive} and \texttt{XLAM} along with the training splits of the \texttt{CustomTools} and \texttt{IslamicRAGTool} datasets. 
\end{itemize}
In Experiments 3–5, we use the instruction-tuned Fanar model that differs from the base pre-trained model used in Experiments~1 and~2. This model has undergone both supervised fine-tuning and preference learning in Arabic and English, allowing it to more effectively follow user intent across both languages \cite{fanarllm2025}.
}

\subsection{Fine-Tuning Setup}
We fine-tuned all models using supervised learning with LLaMA-Factory~\cite{zheng2024llamafactory}.
The training setup is the same for all models: we use a cosine learning rate schedule with a peak learning rate of $5.0 \times 10^{-7}$ and a minimum of $5.0 \times 10^{-8}$, and a batch size of 640.
We fine-tune two public models: Fanar-1-9B, a pre-trained base model, and Fanar-1-9B-Instruct, its post-trained variant~\cite{fanarllm2025} to measure the effect of SFT on tool calling capabilities.

\subsection{Evaluation Methodology}

We fine-tuned the models to produce one of two outputs: a dedicated \texttt{<no\_tool\_call>} tag when no action is required, or a function call, with tool name and arguments, encapsulated within \texttt{<tool\_call></tool\_call>} tags.  For evaluation, each model is tested on all test splits detailed in Table~\ref{tab:datasets}. To ensure a fair comparison with single-turn datasets, we decompose the multi-turn conversations from the \texttt{Glaive} test set into individual turns. We report the weighted-average precision and recall across all available tools, where the weighting reflects the relative importance of each tool based on its frequency in the test set.

\asim{Our evaluation methodology employs two complementary approaches: function name detection and end-to-end argument accuracy. First, we calculate the precision ($P_T$) and the recall ($R_T$) for each tool $T$ \asim{based on function name matching only.} For each tool/class, precision measures the fraction of predicted tool calls that are correct, while recall measures the fraction of actual tool calls that are correctly identified. Notably, we treat the absence of a tool call as its own tool, representing cases where no function tool is invoked:}

\begin{align*}
    P_T &= \frac{\text{True Positives}_T}{\text{True Positives}_T + \text{False Positives}_T} \\
    R_T &= \frac{\text{True Positives}_T}{\text{True Positives}_T + \text{False Negatives}_T}
\end{align*}
These individual scores are then aggregated using a weighted average, where each tool's contribution is weighted by its support ($N_T$)—the number of true instances in the test set. The final weighted-average metrics are defined as:
\begin{align*}
    \text{Precision}_{\text{weighted}} &= \sum_{T \in K} \frac{N_T}{N_{\text{total}}} \cdot P_T \\
    \text{Recall}_{\text{weighted}} &= \sum_{T \in K} \frac{N_T}{N_{\text{total}}} \cdot R_T
\end{align*}
where $K$ is the set of all tools and $N_{\text{total}}$ is the total number of instances.

\asim{Beyond function name detection, we assess end-to-end performance through Argument Population Accuracy (ArgA), which quantifies the proportion of function calls where both the function name and all parameter values are correctly predicted. This comprehensive metric evaluates the model's capacity to not only select the appropriate tool but also furnish it with accurate argument values:
\begin{align*}
    \text{ArgA} &= \frac{\text{Exact Matches}}{\text{Total Positive Cases}}
\end{align*}
where Exact Matches denotes instances with perfect correspondence in both function name and arguments, and Total Positive Cases encompasses all cases requiring function calls (excluding \texttt{<no\_tool\_call>} instances). ArgA delivers a holistic evaluation of the model's practical effectiveness in real-world function calling applications.}

\asim{To ensure reliable ArgA computation, we implement standardized normalization protocols for both ground truth and predicted function calls prior to assessment. These normalizations include lowercase normalization, elimination of extraneous whitespaces, and standardization of date formats and numerical representations. This preprocessing is essential because models may generate semantically identical outputs with minor formatting discrepancies (e.g., ``2024-01-15'' versus ``2024/01/15'' for dates, or ``John Smith'' versus ``john smith''). By applying uniform normalization rules to both reference and predicted outputs, we focus  evaluation on semantic accuracy rather than superficial formatting differences, yielding a more precise assessment of functional performance.}

\begin{table*}
\centering
\setlength{\tabcolsep}{2.3pt} 
\small
\caption{\asim{Performance evaluation across five training configurations showing precision (P) and recall (R) for the function call detection task (measuring whether function names match), and argument population accuracy (ArgA) for end-to-end correctness requiring both correct function names and argument values. Training setups: (1) English-only tool-calling data, trained with a random mix of Glaive EN and xLAM EN; (2) Arabic-only tool-calling data, trained with a random mix of Glaive AR and xLAM AR; (3) Supervised Fine-Tuning (SFT) followed by training on a random mix of Glaive EN and xLAM EN; (4) SFT followed by a bilingual (EN + AR) random mix of Glaive and xLAM; (5) SFT followed by a bilingual (EN + AR) random mix of Glaive and xLAM, IslamicRAGTool and CustomTools. Test sets are evaluated in Arabic (AR) and English (EN). Function Calling (FC) indicates whether the test set contains positive cases requiring function calls (Yes) or negative cases without function calls (No).}}

\label{tab:results}
\begin{tabular}{lllccccccccccccccc}
\hline
\textbf{Dataset} & Language & FC & \multicolumn{3}{c}{\textbf{Exp. 1}} & \multicolumn{3}{c}{\textbf{Exp. 2}} & \multicolumn{3}{c}{\textbf{Exp. 3}} & \multicolumn{3}{c}{\textbf{Exp. 4}} & \multicolumn{3}{c}{\textbf{Exp. 5}} \\
\cline{4-18}
& & & P & R & ArgA & P & R & ArgA & P & R & ArgA & P & R & ArgA & P & R & ArgA \\
\hline
\texttt{Glaive} & AR & Yes & 1.00 & 0.99 & 0.69	 & 1.00 & 0.99 & 0.78 & 1.00 & 1.00 & 0.71 & 1.00 & 0.99 & 0.77 & 1.00 & 0.99 & 0.77 \\
& & No & 1.00 & 0.95 & - & 1.00 & 0.98 & - & 1.00 & 0.96 & - & 1.00 & 0.99 & - & 1.00 & 1.00 & - \\
\cline{4-18}
& EN & Yes & 1.00 & 0.99 & 0.90 & 1.00 & 0.99 & 0.88 & 1.00 & 0.99 & 0.91 & 1.00 & 0.99 & 0.91 & 1.00 & 0.99 & 0.91 \\

& & No & 1.00 & 0.99 & - & 1.00 & 0.98 & - & 1.00 & 0.99 & - & 1.00 & 0.99 & - & 1.00 & 0.99 & - \\
\hline
\texttt{xLAM} & AR & Yes & 0.97 & 0.97 & 0.61 & 0.98 & 0.98 & 0.75 &  0.98 & 0.98 &  0.62 & 0.99 & 0.98 & 0.76 & 0.98 & 0.98 & 0.76 \\
& & No & 1.00 & 0.98 & - & 1.00 & 0.98 & - & 1.00 & 0.97 & - & 1.00 & 0.99 & - & 1.00 & 0.99 & - \\
\cline{4-18}
& EN & Yes & 0.98 & 0.98 & 0.85 & 0.98 & 0.99 & 0.82 & 0.98 & 0.98 & 0.86 & 0.98 & 0.98 & 0.87 & 0.99 & 0.99 & 0.86 \\
& & No & 1.00 & 0.98 & - & 1.00 & 0.97 & - & 1.00 & 0.98 & - & 1.00 & 0.99 & - & 1.00 & 0.99 & - \\
\hline
\texttt{CustomTools} & AR & Yes & 0.98 & 0.66 & 0.45 & 0.97 & 0.82 & 0.77 & 0.98 & 0.86 & 0.58 & 0.98 & 0.86 & 0.80 & 1.00 & 1.00 & 1.00 \\
& & No & 1.00 & 0.97 & - & 1.00 & 0.90 & - & 1.00 & 0.74 & - & 1.00 & 0.89 & - & 1.00 & 1.00 & - \\
\cline{4-18}
& EN & Yes & 0.97 & 0.70 & 0.56 & 0.96 & 0.80 &0.56	 & 0.96 & 0.80 & 0.64 & 0.96 & 0.81 & 0.63 & 1.00 & 0.99 & 1.00 \\
& & No & 1.00 & 0.98 & - & 1.00 & 0.92 & - & 1.00 & 0.87 & - & 1.00 & 0.94 & - & 1.00 & 1.00 & - \\
\hline
\texttt{IslamicRAGTool} & AR & Yes & 1.00 & 0.25 & 0.14 & 1.00 & 0.47 & 0.36 & 1.00 & 0.69 & 0.42 & 1.00 & 0.63 & 0.49 & 1.00 & 0.99 &  0.99\\
& & No & 1.00 & 0.98 & - & 1.00 & 0.94 & - & 1.00 & 0.90 & -& 1.00 & 0.95 & - & 1.00 & 1.00 & - \\
\cline{4-18}
& EN & Yes & 1.00 & 0.44 & 0.33	& 1.00 & 0.58 & 0.33& 1.00 & 0.71 &0.54 & 1.00 & 0.62 & 0.51 & 1.00 & 0.99& 0.99	  \\
& & No & 1.00 & 0.97 & - & 1.00 & 0.95 & - & 1.00 & 0.95 & - & 1.00 & 0.95 & - & 1.00 & 1.00 & - \\
\hline
\end{tabular}
\end{table*}

\section{Results and Analysis}

Table \ref{tab:results} presents the comprehensive results of all the experiments conducted. As expected, models achieve nearly perfect precision and recall when evaluated on test examples drawn from the same domain as the training data. This pattern is consistently observed across the Glaive and xLAM test sets, where all models were trained on the respective training portions of these datasets, regardless of whether they used Arabic, English, or bilingual training data.
To address our three research questions, we turn our attention to the cross-domain evaluation results obtained from the remaining datasets, which provide insights into the models' generalization capabilities beyond their training domains. 

\subsection{Cross-Lingual Knowledge Transfer in Tool Calling}
\taha{We examine the transferability of tool-calling capabilities between English and Arabic by comparing the results of Experiment~1 and Experiment~2. The results indicate that models trained on tool-calling data in one language (English or Arabic) can effectively transfer this ability to the other language. This suggests that the base model’s translation capabilities are sufficiently robust to cross-lingually detect the correct tool calls.
However, when evaluating on previously unseen tools, particularly domain-specific ones such as \texttt{CustomTools} and \texttt{IslamicRAGTool}, we observe a significant drop in recall, where the LLM should have invoked a tool but didn't. This gap becomes more pronounced when moving from moderately custom tools (e.g., 0.66-0.82 for \texttt{CustomTools}) to highly specialized ones (e.g., 0.25-0.47 \texttt{IslamicRAGTool}). Notably, this decline occurs regardless of the training language (either Arabic, English, or both). 
This highlights a broader generalization gap in tool invocation for previously unseen tools, especially those with niche or specialized behavior.} 
\taha{
Interestingly, we find that training on Arabic tool-calling data yields slightly better generalization to English than the reverse, with a consistent performance gap of approximately 0.1–0.2, depending on the dataset. This asymmetry may stem from the domain-specific nature of the custom tools, which are more richly represented in the Arabic fine-tuning datasets. As a result, the model benefits from exposure to these specialized contexts during training, which in turn enhances its ability to generalize to English inputs.
}

\asim{As for argument population accuracy (ArgA), the results show that a mismatch in the language of training versus testing data adversely affects the ability of the model to guess the correct arguments, particularly for unseen tools.  For example, ArgA dropped for the Arabic test set from 0.78 to 0.69 and from 0.75 to 0.61 for Glaive and xLAM respectively when training with English versus Arabic.  An even sharper decline was observed for \texttt{CustomTools} and \texttt{IslamicRAGTool} with a drop from 0.77 to 0.45 and from 0.36 to 0.14 respectively. 
This underscoring that the model struggles not only with deciding when to call a tool, but also with correctly populating its arguments.}

\subsection{In-Language Fine-Tuning}

The addition of Arabic tool-calling data to the English fine-tuning dataset (transitioning from Experiment 3 to Experiment 4) produces notable improvements in non-function-calling performance. For the \texttt{CustomTools} dataset, non-FC recall increases substantially from 0.74 to 0.89 for Arabic test sets and from 0.87 to 0.94 for English test sets. Low non-FC recall indicates that the LLM chose a wrong tool instead of returning <no\_tool\_call>. In contrast, function-calling cases show minimal improvement, with English recall increasing slightly from 0.80 to 0.81 while Arabic recall remains unchanged. The \texttt{IslamicRAGTool} dataset exhibits a similar pattern for non-FC cases, demonstrating consistent benefits from bilingual training data. However, an unexpected trend emerges in the FC cases, where performance actually decreases. This declining pattern is not isolated to \texttt{IslamicRAGTool} but occurs across approximately half of the individual tools within \texttt{CustomTools} when comparing Experiments 3 and 4. Despite these localized drops, the overall weighted average recall remains positive, indicating that the benefits of including Arabic data outweigh the drawbacks.
\asim{A more significant trend is visible in argument population accuracy, which improves markedly for Arabic test cases in both \texttt{CustomTools} (from 0.58 to 0.80) and \texttt{IslamicRAGTool} (from 0.42 to 0.49), while slightly decreasing for the corresponding English cases.} 

\subsection{Effect of General SFT}

The effect of general SFT data is most evident when comparing Experiment 1 and Experiment 3, revealing contrasting impacts on function-calling (FC) and non-function-calling cases across different datasets. For function-calling cases, the General SFT data produces substantial improvements in recall performance. In the \texttt{CustomTools} dataset, recall increases significantly from 0.66 to 0.86 for Arabic and from 0.70 to 0.80 for English, \asim{with argument population accuracy also rising from 0.45 to 0.58 and 0.56 to 0.64, respectively.} The improvements are even more pronounced in the \texttt{IslamicRAGTool} dataset, where Arabic recall jumps from 0.25 to 0.69 and English recall rises from 0.44 to 0.71, \asim{accompanied by a major boost in ArgA from 0.14 to 0.42 for Arabic and 0.33 to 0.54 for English.}

However, non-function-calling cases show a concerning decline in performance after applying general SFT data. The \texttt{CustomTools} dataset experiences notable drops in recall, falling from 0.97 to 0.74 in Arabic and from 0.98 to 0.87 in English. The \texttt{IslamicRAGTool} dataset shows a more modest decline, with Arabic recall dropping from 0.98 to 0.90 and English recall decreasing from 0.97 to 0.95. 
\taha{The performance decline in non-function-calling (non-FC) cases is likely due to the supervised fine-tuning (SFT) data enhancing the model’s generative abilities while diminishing its classification precision, leading the model to incorrectly predict function calls in cases where none are required.}
This suggests that the general training data may be introducing a bias toward function-calling behavior.

Notably, Experiment 4 demonstrates that adding Arabic tool-calling data can help recover some of the lost performance. The Arabic non-FC recall improves from 0.74 to 0.89, indicating that language-specific training data can help balance the model's classification behavior and mitigate the negative effects of overly confident function-calling predictions. 

\subsection{Tool-Specific Fine-Tuning}
To address whether fine-tuning LLMs on tool-specific data is necessary, Experiment 5 involves training on all available datasets simultaneously. This comprehensive approach accounts for the substantial performance gains observed when comparing Experiment 5 to all previous experiments. The \texttt{CustomTools} and \texttt{IslamicRAGTool} results exemplify this improvement, \asim{with both recall and argument population accuracy scores reaching 0.99 or higher in most cases.} These results demonstrate the effectiveness of fine-tuning on task-specific data, effectively eliminating classification \asim{and agrument population} errors. 
This behavior aligns with the fact that training on Glaive and XLAM data yields nearly perfect tool selection results when tested on their respective test splits.

To test generalization, we tested a publicly available multilingual LLM, namely the instruction-tuned Qwen2.5-7B \cite{qwen2.5}, which was tuned for tool-calling and is comparable in size to Fanar 9B, with and without additional instruction tuning using the training splits of \texttt{CustomTools} and \texttt{IslamRAGTool}.  We tested on the \texttt{CustomTools} and \texttt{IslamRAGTool} only, because we cannot exclude the possibility that Qwen was trained Glaive and/or xLAM. Table \ref{tab:qwen_comparison_pr_single} shows the Qwen results with and without additional finetuning (FT and Base respectively).  The results show that additional finetuning for the tools of interest yields a very large boost in tool-calling effectiveness, \asim{with both recall and argument population accuracy showing dramatic improvement.} For example, the recall for \texttt{IslamicRAGTool} for English when tool-calling was required improved from 0.66 to 0.91, \asim{ while ArgA jumped from 0.46 to 0.91.} Nonetheless, the results of Qwen with continued finetuning falls short of the best Fanar results (Experiment 5), particularly for \texttt{IslamicRAGTool}.  We suspect that this is the result of Fanar being specifically pre-trained on Arabic and Islamic content. 

\begin{table}[htbp!]
\centering
\setlength{\tabcolsep}{3.5pt}
\scriptsize 

\caption{\asim{Performance comparison of the base Qwen2.5-7B-Instruct model versus a version fine-tuned on the CustomTools and IslamicRagTool datasets. Metrics reported are precision (P) and recall (R) for the function call detection task (measuring whether function names match), and argument population accuracy (ArgA) for end-to-end correctness requiring both correct function names and argument values. Test sets are evaluated in Arabic (AR) and English (EN). Function Calling (FC) indicates whether the test set contains positive cases requiring function calls (Yes) or negative cases without function calls (No).}}

\label{tab:qwen_comparison_pr_single}
\begin{tabular}{@{}llccccccc@{}}
\toprule
\multirow{2}{*}{\textbf{Dataset}} & \multirow{2}{*}{\textbf{Language}} & \multirow{3}{*}{\textbf{FC}} & \multicolumn{3}{c}{\textbf{Base}} & \multicolumn{3}{c}{\textbf{FT}} \\
\cmidrule{4-9}
& & & \textbf{P} & \textbf{R} & \textbf{ArgA} & \textbf{P} & \textbf{R} & \textbf{ArgA} \\
\midrule
\multirow{4}{*}{CustomTools} & \multirow{2}{*}{AR} & Y & 0.95 & 0.85 & 0.64 & 0.99 & 0.96 & 0.95 \\
& & N & 1.00 & 0.71 & - & 1.00 & 1.00 & - \\
\cmidrule(l){4-9}
& \multirow{2}{*}{EN} & Y & 0.97 & 0.94  & 0.74 & 0.99 & 0.98 & 0.98 \\
& & N & 1.00 & 0.81 & - & 1.00 & 1.00 & - \\
\midrule
\multirow{4}{*}{IslamicRAGTool} & \multirow{2}{*}{AR} & Y & 1.00 & 0.70 & 0.45	& 1.00 & 0.92 & 0.91 \\
& & N & 1.00 & 0.89 & - & 1.00 & 1.00 & - \\
\cmidrule(l){4-9}
& \multirow{2}{*}{EN} & Y & 1.00 & 0.66 & 0.46 & 1.00 & 0.91 & 0.91 \\
& & N & 1.00 & 0.94 & - & 1.00 & 1.00 & - \\
\bottomrule
\end{tabular}
\end{table}

\subsection{Deeper Analysis of Argument Population Accuracy}

\asim{While precision and recall measure a model's ability to \textit{select} the correct tool, the Argument Population Accuracy (\texttt{ArgA}) metric evaluates the more challenging task of end-to-end correctness, requiring both the function name and all argument values to be perfect. Across all experiments, a significant gap exists between tool-calling recall and the corresponding \texttt{ArgA} score, indicating that correctly populating arguments is a primary bottleneck for performance.}

\asim{To understand the sources of ArgA failures, we conducted a detailed error analysis focusing on cases where function names were correctly identified but argument values were erroneous. From a total of 7,211 errors from all the experiments, we randomly selected 249 cases (see breakdown in Appendix~\ref{sec:ArgStats}) and systematically categorized them as follows: 
\begin{itemize}[leftmargin=*]
    \item \textbf{W} (Wrong argument values): The model produced incorrect arguments; 
    \item \textbf{T} (Translation discrepancy): Expected argument values are in one language while model outputs are in another;
    \item \textbf{P} (Paraphrasing variance): Predicted arguments are paraphrases of the expected ones;
    \item \textbf{I} (Incomplete context): The query originates from a multi-turn conversation and lacks essential information, rendering certain argument values unpopulatable without prior conversational context.
\end{itemize}
}

\asim{The most frequent error category was Paraphrasing Variance (P), accounting for 50.2\% of all argument errors. This occurs when the model generates a semantically correct argument that is syntactically different from the ground truth (e.g., "Could you tell me what Islam is?" vs. "What is Islam?"). This error type was particularly dominant in the English test sets (73.6\% of errors) and more so in the specialized \texttt{IslamicRAGTool} dataset (82.7\% of errors). 
This finding directly explains the dramatic success of Experiment 5, where tool-specific fine-tuning on all datasets resulted in near-perfect \texttt{ArgA} scores. Training on exact target examples, the model learns the exact syntactic format required, effectively eliminating paraphrasing ambiguities.}

\asim{The second most common issue was Translation Discrepancy (T), making up 38.2\% of errors. This error was overwhelmingly concentrated in the Arabic test sets, where it was the leading cause of failure (53.1\% of all Arabic errors). This insight is critical for interpreting the cross-lingual experiments. The low \texttt{ArgA} scores in Experiment 1, where an English-trained model was tested on Arabic, can be directly attributed to the model's tendency to provide arguments in English instead of Arabic. In contrast, a significant improvement in \texttt{ArgA} when bilingual data was introduced in Experiment 4 (e.g., increasing from 0.58 to 0.80 for \texttt{CustomTools} AR) demonstrates that bilingual fine-tuning is essential for teaching the model the correct language of the expected argument.}

\asim{Finally, Wrong Argument Values (W) and Incomplete Context (I) were less frequent (6.8\% and 4.8\%, respectively). The latter category refers to cases where the user's query originates from a multi-turn conversation and lacks essential information from previous turns, making it impossible to populate certain arguments.}

\asim{In summary, this deeper analysis reveals that the primary obstacles to achieving high end-to-end tool-calling accuracy are not necessarily comprehension, but rather adherence to specific formatting rules. Cross-lingual performance is hindered by a failure to translate arguments, while generalization to new tools is challenged by syntactic ambiguity. These findings suggest that such errors are best mitigated by reducing ambiguity. The most effective approach, demonstrated in our experiments, is providing direct, in-domain examples through tool-specific fine-tuning. An alternative would be to craft highly granular function and argument descriptions. By explicitly defining expected formats, such as date conventions or required languages, such descriptions could guide the model's behavior through in-context learning, potentially reducing the need for extensive fine-tuning data.}


\section{Conclusion}
\taha{We conducted a series of experiments to investigate how tool-calling performance is influenced by language and the ability to generalize to previously unseen tools. Our findings highlight the importance of training on bilingual datasets, performing instruction tuning, and providing explicit examples of tool usage during fine-tuning. Most notably, we find that when developing agentic frameworks tailored to specific custom tools, direct fine-tuning on those tools is significantly more effective compared to relying on generalization alone. In practice, this may entail continued fine-tuning of an instruction tuned model that is capable of tool calling with training examples for the tools of interest.}

\section*{Limitations}
\taha{Our conclusions are primarily based on experiments using two stock datasets—Glaive and xLAM—which may not fully capture the diversity of tool-calling use cases, especially in domain-specific or low-resource settings. While these datasets provide valuable benchmarks, extending the analysis to additional datasets could offer a more comprehensive view of language and generalization effects. Furthermore, our evaluation focuses on recall-based metrics and does not account for downstream utility or correctness of tool execution in real-world agentic systems. Finally, although we consider English and Arabic, additional languages with different morphological and syntactic properties may exhibit different transfer dynamics, warranting further investigation.}

\bibliography{custom}

\appendix

\section{Prompts}
\label{sec:prompts}

\begin{tcolorbox}[
  width=\linewidth,
  colback=gray!10,
  colframe=gray!50!black,
  title=System Prompt for Translating Datasets (ex. Glaive),
  breakable,
  enhanced,
  sharp corners,
  boxrule=0.5pt
]
\ttfamily
You are a helpful assistant tasked with translating user queries and function argument values into \textbf{Arabic}, using the descriptions of the tools and their arguments as guidance.

\subsection*{Translation Guidelines}

\begin{itemize}
    \item \textbf{Do not} translate function names or argument keys — only translate the \textbf{values} inside the arguments.
    \item \textbf{Do not} modify any values that are clearly identifiers.
    \item Preserve the original \textbf{JSON structure} exactly as it is.
\end{itemize}

\subsection*{Expected Output Format}

Always return a JSON object with the following two keys:

\begin{itemize}
    \item \texttt{"question"}: The user query, translated into Arabic.
    \item \texttt{"function\_calls"}: A list of function call objects, where only the argument values are translated into Arabic.
\end{itemize}

\end{tcolorbox}

\begin{tcolorbox}[
  width=\linewidth,
  colback=gray!10,
  colframe=gray!50!black,
  title=System Prompts for Synthetic Data Generation,
  breakable,
  enhanced,
  sharp corners,
  boxrule=0.5pt
]
\ttfamily

We do this process in a couple of steps to have the LLM focusing on a few tasks at a time, which yielded better data in our case: \\

\textbf{\# Step 1:} Generating questions that can be answered by the given tools. \\

Generate $N$ unique English example questions that can be answered using the tools listed below. Ensure the following conditions are met:

\begin{itemize}
  \item Each question should be \textbf{distinct}—no repetitions or near-duplicates.
  \item Distribute the questions \textbf{equally} across all tools.
  \item Format the output as a list of \textbf{tuples}, where each tuple contains:
  \begin{itemize}
    \item The question as a string.
    \item The corresponding tool/function name as a string.
  \end{itemize}
\end{itemize}

The output should be a tuple containing the question and the related tool: (query, function-name). \\

\textbf{\# Step 2:} Populate the related arguments of the function given query/function-name pairs. \\

You are given a list of tuples, where each tuple contains:
\begin{itemize}
  \item A \textbf{natural language query} (i.e., a question or instruction), and
  \item The corresponding \textbf{function name} (i.e., the tool that should be invoked to fulfill the query).
\end{itemize}

Each function has a predefined schema specifying its \textbf{required arguments}.

Your task is to extract and populate all \textbf{required argument values} for that function based on the information provided in the query.

For each input tuple, return a new tuple in the following format: (query, function-name, arguments-dict) \\

\textbf{\# Step 3:} Generate an answer and a paraphrase for each answer for every query based on the given function schema \\

Generate a sample answer and a paraphrase of that answer based on the provided question and the corresponding function.

\begin{itemize}
    \item Separate the \textbf{answer} and the \textbf{paraphrase} using a \textbf{double newline character}.
    \item Both the answer and the paraphrase must be written in \textbf{English}.
    \item \textbf{Do not mention or reference the tool or function} used in either the answer or the paraphrase.
    \item Assume that all tools operate correctly and return accurate, expected results without any errors or failures.
    \item \textbf{Do not generate or execute any code}; your output should only be natural language text.
    \item The paraphrase must \textbf{include the answer itself}, but rephrased with different wording and sentence structure.
\end{itemize}

\end{tcolorbox}

\section{Dataset Examples}
\label{sec:datasetExamples}

The following are examples from our translated xLAM and CustomTools datasets. For clarity and readability, we’ve formatted them for presentation. In the original data, however, the tools and function calls are represented in raw JSON format, including detailed information about their arguments.

\begin{tcolorbox}[
  width=\linewidth,
  colback=gray!10,
  colframe=gray!50!black,
  title=Examaple 1: xLAM Translated Example,
  breakable,
  enhanced,
  sharp corners,
  boxrule=0.5pt
]

\textbf{System:} \\
You are a helpful assistant with access to the following functions. Use them if required; otherwise, return no-tool-call. If you need more clarification to be able to use a tool, ask for clarification.  \\

\vspace{0.5em}
Available Tools:
\begin{itemize}
  \item \texttt{sort\_numbers(numbers, descending)}: Sorts a list of numbers.
  \item \texttt{binomial\_probability(n, k, p)}: Calculates binomial probability.
  \item \texttt{wire\_resistance(length\_m, area\_sq\_m, material)}: Calculates the resistance of a wire.
\end{itemize}

\vspace{1em}

\textbf{User:} \\
\begin{otherlanguage}{arabic}
قم بترتيب الأرقام [3.2، 1.5، 2.7، 4.1] بترتيب تصاعدي واحسب احتمال ثنائي الحصول على 3 نجاحات في 5 تجارب مع احتمال نجاح 0.6.
\end{otherlanguage}

\vspace{1em}

\textbf{Assistant:} \\
Tool Calls:
\begin{itemize}
  \item sort\_numbers(numbers=[3.2, 1.5, 2.7, 4.1], descending=false)
  \item binomial\_probability(n=5, k=3, p=0.6)
\end{itemize}

\end{tcolorbox}

\begin{tcolorbox}[
  width=\linewidth,
  colback=gray!10,
  colframe=gray!50!black,
  title=Examaple 2: CustomTools Arabic Example,
  breakable,
  enhanced,
  sharp corners,
  boxrule=0.5pt
]

\textbf{System:} \\
You are a helpful assistant with access to the following functions. Use them if required; otherwise, return no-tool-call. If you need more clarification to be able to use a tool, ask for clarification.  \\

\vspace{0.5em}
Available Tools:
\begin{itemize}
  \item \texttt{generate\_image(prompt, size)}: Generate an image using an image generation API.
  \item \texttt{generate\_speech(text, save\_path, voice)}: Convert written text (Arabic or English) to speech.
  \item \texttt{islamic\_knowledge(question)}: Use for questions specifically about Islam (\foreignlanguage{arabic}{الإسلام}), including beliefs Aqeedah (\foreignlanguage{arabic}{العقيدة}), Quran (\foreignlanguage{arabic}{القرآن}), Hadith (\foreignlanguage{arabic}{الحديث}), Sunnah (\foreignlanguage{arabic}{السنة}), Fiqh (\foreignlanguage{arabic}{الفقه}), Seerah (\foreignlanguage{arabic}{السيرة}), Islamic history, rulings, and practices. \\
  \item \texttt{recent\_news(question)}: Answer questions about recent events or current affairs.
  \item \texttt{diacritize\_text(text, process\_for\_tts)}: Add Arabic diacritics (\foreignlanguage{arabic}{تشكيل}) to text.
  \item \texttt{person\_biography\_lookup(question)}: Answer biographical questions.
  \item \texttt{translate\_text(text, source\_lang, target\_lang)}: Translate text between languages.
  \item \texttt{transcribe\_audio(audio\_file\_path)}: Transcribe audio files using ASR.
\end{itemize}

\vspace{1em}

\textbf{User:} \\
\foreignlanguage{arabic}{ما الفرق بين الإلهام والرؤيا الصالحة؟}

\vspace{1em}

\textbf{Assistant:} \\
Tool Calls:
\begin{itemize}
  \item {islamic\_knowledge(question=\foreignlanguage{arabic}{ما الفرق بين الإلهام والرؤيا الصالحة؟}})
\end{itemize}

\vspace{1em}

\textbf{Tool Response:} \\
\foreignlanguage{arabic}{الفرق بين الإلهام والرؤيا الصالحة يكمن في مصدرهما وكيفية تلقيهما. الإلهام هو ما يلقيه الله في القلب من معانٍ وأفكار صالحة، ويكون خفيًا لا يدرك مصدره الإنسان تحديدًا، وقد يكون في اليقظة أو المنام. أما الرؤيا الصالحة فهي جزء من ستة وأربعين جزءًا من النبوة، وتكون في المنام فقط، وهي بشارة من الله أو إنذار أو تعليم. الإلهام قد يكون لأي مؤمن، بينما الرؤيا الصالحة غالبًا ما تكون للمؤمنين الصادقين.}

\vspace{1em}

\textbf{Assistant:} \\
\foreignlanguage{arabic}{يُقال أن الفرق بين الإلهام والرؤيا الصالحة هو في منشئهما وكيفية إدراكهما، حيث أن الإلهام يأتي من الله مباشرة إلى القلب بشكل غير مباشر وواضح، في حين أن الرؤيا الصالحة تأتي في المنام وتعتبر جزءًا من النبوة.}

\end{tcolorbox}

\section{Custom Tools Definitions} 
\label{sec:customToolsDef}
We list below the definitions of the tools found in our CustomTools Dataset.

\begin{description}

  \item[\texttt{translate\_text}] \hfill \\
  \textbf{Description:} Translates text from one language to another using a reliable and efficient translation API. \\
  \textbf{Parameters:}
  \begin{itemize}
    \item \texttt{text} (string): The text to be translated.
    \item \texttt{source\_lang} (string): The source language code. e.g. en, ar, etc.
    \item \texttt{target\_lang} (string): The target language code.
  \end{itemize}

  \item[\texttt{generate\_speech}] \hfill \\
  \textbf{Description:} Text-to-Speech (TTS) converts written text in either English or Arabic to speech. \\
  \textbf{Parameters:}
  \begin{itemize}
    \item \texttt{text} (string): The text to convert to speech.
    \item \texttt{save\_path} (string): Path to save the audio file. If None, creates a default path.
    \item \texttt{voice} (string): The voice to use. Defaults to 'default'.
  \end{itemize}

  \item[\texttt{generate\_image}] \hfill \\
  \textbf{Description:} Generate an image using an image generation API. \\
  \textbf{Parameters:}
  \begin{itemize}
    \item \texttt{prompt} (string): Description of the image to generate.
    \item \texttt{size} (string): Image size. Defaults to '1024x1024'.
  \end{itemize}

  \item[\texttt{islamic\_knowledge}] \hfill \\
  \textbf{Description:} Use for questions specifically about Islam (\foreignlanguage{arabic}{الإسلام}), including beliefs Aqeedah (\foreignlanguage{arabic}{العقيدة}), Quran (\foreignlanguage{arabic}{القرآن}), Hadith (\foreignlanguage{arabic}{الحديث}), Sunnah (\foreignlanguage{arabic}{السنة}), Fiqh (\foreignlanguage{arabic}{الفقه}), Seerah (\foreignlanguage{arabic}{السيرة}), Islamic history, rulings, and practices. \\
  \textbf{Parameters:}
  \begin{itemize}
    \item \texttt{question} (string): The question about Islamic knowledge or teachings.
  \end{itemize}

  \item[\texttt{transcribe\_audio}] \hfill \\
  \textbf{Description:} Transcribe audio using an ASR API. \\
  \textbf{Parameters:}
  \begin{itemize}
    \item \texttt{audio\_file\_path} (string): Path to the audio file.
  \end{itemize}

  \item[\texttt{person\_biography\_lookup}] \hfill \\
  \textbf{Description:} Answer biographical question about a person. \\
  \textbf{Parameters:}
  \begin{itemize}
    \item \texttt{question} (string): The biographical question.
  \end{itemize}

  \item[\texttt{diacritize\_text}] \hfill \\
  \textbf{Description:} Adds Arabic diacritics (tashkeel – \foreignlanguage{arabic}{تشكيل}) to Arabic text. \\
  \textbf{Parameters:}
  \begin{itemize}
    \item \texttt{text} (string): Arabic text to diacritize.
    \item \texttt{process\_for\_tts} (boolean): Whether to optimize diacritization for text-to-speech. Defaults to False.
  \end{itemize}

  \item[\texttt{recent\_news}] \hfill \\
  \textbf{Description:} Answer questions about recent events, news, and current affairs. \\
  \textbf{Parameters:}
  \begin{itemize}
    \item \texttt{question} (string): The question about recent events or information.
  \end{itemize}

\end{description}

\section{Detailed Breakdown of ArgA Error Categories}
\label{sec:ArgStats}

Table~\ref{tab:error_analysis} shows the distribution of \texttt{ArgA} error categories across the five experiments, separated by Arabic and English.

\begin{table}[h]
\centering
\scriptsize
\setlength{\tabcolsep}{2.5pt} 
\begin{tabular}{@{}p{1.0cm}c p{1.2cm} p{1.2cm} c c c@{}}
\toprule
\multicolumn{6}{c}{\textbf{Arabic (AR)}} \\
\midrule
\textbf{Exp} & \textbf{Total} & \textbf{P} & \textbf{T} & \textbf{W} & \textbf{I} \\
\midrule
\textbf{Exp. 1} & 33 & 7 (21.2\%) & \textbf{24 (72.7\%)} & 1 (3.0\%) & 1 (3.0\%)\\
\textbf{Exp. 2} & 29 & 12 (41.4\%) & \textbf{13 (44.8\%)} & 3 (10.3\%) & 1 (3.4\%)  \\
\textbf{Exp. 3} & 44 & 15 (34.1\%) & \textbf{25 (56.8\%)} & 0 (0.0\%) & 4 (9.1\%) \\
\textbf{Exp. 4} & 33 & \textbf{17 (51.5\%)} & 13 (39.4\%) & 3 (9.1\%) & 0 (0.0\%)  \\
\textbf{Exp. 5} & 23 & 10 (43.5\%) & \textbf{11 (47.8\%)} & 1 (4.3\%) & 1 (4.3\%) \\
\midrule
\textbf{Total} & \textbf{162} & 61 (37.7\%) & \textbf{86 (53.1\%)} & 8 (4.9\%) & 7 (4.3\%) \\
\midrule
\\[-0.2cm] 
\multicolumn{6}{c}{\textbf{English (EN)}} \\
\midrule
\textbf{Exp} & \textbf{Total} & \textbf{P} & \textbf{T} & \textbf{W} & \textbf{I}  \\
\midrule
\textbf{Exp. 1} & 16 & \textbf{14 (87.5\%)} & 0 (0.0\%) & 1 (6.2\%) & 1 (6.2\%) \\
\textbf{Exp. 2} & 26 & \textbf{13 (50.0\%)} & 9 (34.6\%) & 3 (11.5\%) & 1 (3.8\%) \\
\textbf{Exp. 3} & 23 & \textbf{18 (78.3\%)} & 0 (0.0\%) & 4 (17.4\%) & 1 (4.3\%)  \\
\textbf{Exp. 4} & 16 & \textbf{14 (87.5\%)} & 0 (0.0\%) & 0 (0.0\%) & 2 (12.5\%)  \\
\textbf{Exp. 5} & 6 & \textbf{5 (83.3\%)} & 0 (0.0\%) & 1 (16.7\%) & 0 (0.0\%) \\
\midrule
\textbf{Total} & \textbf{87} & \textbf{64 (73.6\%)} & 9 (10.3\%) & 9 (10.3\%) & 5 (5.7\%)  \\
\bottomrule
\end{tabular}
\caption{Distribution of \texttt{ArgA} error categories across experiments and languages. Categories: \textbf{P} (paraphrasing variance), \textbf{T} (translation discrepancy), \textbf{W} (wrong values), \textbf{I} (incomplete context). Bolded values mark the most frequent error category}

\label{tab:error_analysis}
\end{table}

\end{document}